\documentclass[letterpaper, 10 pt, conference]{ieeeconf} 

\IEEEoverridecommandlockouts                              

\usepackage{graphics} 
\usepackage{epsfig} 
\usepackage{times} 
\usepackage{amsmath} 
\usepackage{amssymb}  
\usepackage{amsfonts}
\usepackage[colorlinks,linkcolor=black,anchorcolor=black,citecolor=black,urlcolor=black]{hyperref}
\usepackage[table]{xcolor}
\usepackage{booktabs}
\usepackage{multirow}
\usepackage{graphicx}
\usepackage{subfigure}
\usepackage{cite}
\usepackage{float}
\usepackage{makecell}
\usepackage{booktabs}
\usepackage{etoolbox}
\usepackage{bbding}
\usepackage{makecell}
\usepackage{caption}

\makeatletter
\patchcmd{\@makecaption}
  {\scshape}
  {}
  {}
  {}
\makeatletter
\patchcmd{\@makecaption}
  {\\}
  {.\ }
  {}
  {}
\makeatother
\def\tablename{Table}

\graphicspath{{./}}

\title{\LARGE \bf
OmniNxt: A Fully Open-source and Compact Aerial Robot \\ with Omnidirectional Visual Perception
}

\author{Peize Liu$^{1}$, Chen Feng$^{1,\dag}$, Yang Xu$^{2}$, Yan Ning$^{1}$,  Hao Xu$^{1,\dag}$, and Shaojie Shen$^{1}$
\thanks{$^{1}$Department of Electronic and Computer Engineering, The Hong Kong University of Science and Technology, Hong Kong, China.}
\thanks{$^{2}$Division of Emerging Interdisciplinary Areas, The Hong Kong University of Science and Technology, Hong Kong, China.}
\thanks{Email: {\tt\footnotesize {\{pliuan,cfengag,yxuew,yningaa,hxubc\}}@ust.hk},
\tt\footnotesize {eeshaojie@ust.hk}}
\thanks{\textbf{$^{\dag}$ Corresponding Authors}}
}
\begin{document}


\makeatletter
\let\@oldmaketitle\@maketitle
\renewcommand{\@maketitle}{\@oldmaketitle
  \begin{center}
    \centering
    \captionsetup{type=figure}
    \includegraphics[width=0.99\textwidth]{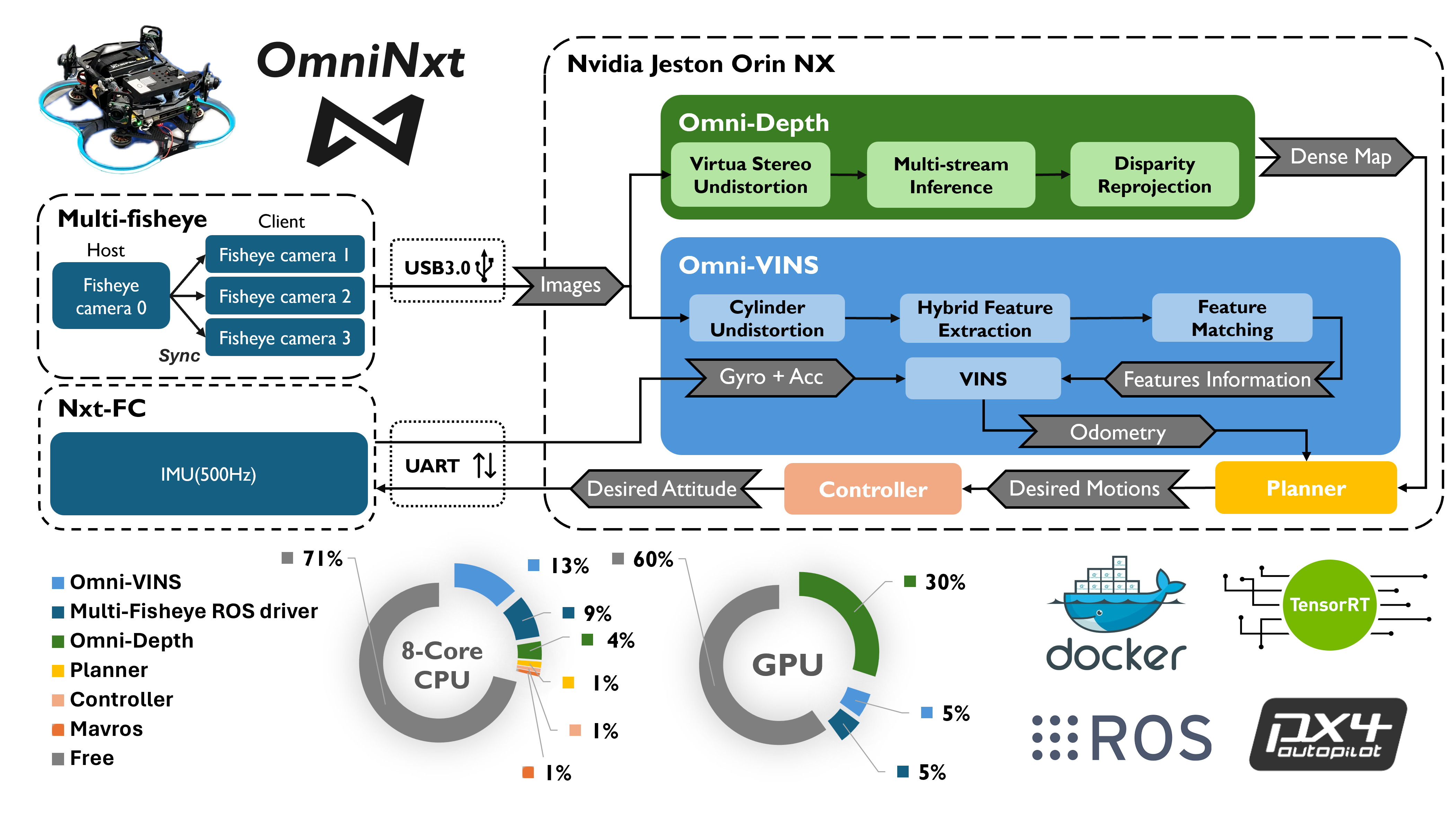}
    \vspace{-0.7cm}
    \captionof{figure}{The system overview of OmniNxt. The hardware architecture includes: Multi-fisheye camera set (camera), Nxt-FC (flight controller and IMU), and Nvidia Jeston Orin NX (onboard computation). The software framework consists of two critical components: Omni-VINS (Sec. \ref{subsec:omnidirectional_vins}) and Omni-Depth (Sec. \ref{subsec:omnidirectional_depth_estimation}).}
    \label{fig:system_overview}
    \vspace{-0.9cm}
\end{center}%
  \bigskip}
\makeatother

\maketitle

\begin{abstract}
Adopting omnidirectional Field of View (FoV) cameras in aerial robots vastly improves perception ability, significantly advancing aerial robotics's capabilities in inspection, reconstruction, and rescue tasks. However, such sensors also elevate system complexity, \textit{e.g.}, hardware design, and corresponding algorithm, which limits researchers from utilizing aerial robots with omnidirectional FoV in their research. To bridge this gap, we propose OmniNxt, a fully open-source aerial robotics platform with omnidirectional perception. We design a high-performance flight controller Nxt-FC and a multi-fisheye camera set for OmniNxt. Meanwhile, the compatible software is carefully devised, which empowers OmniNxt to achieve accurate localization and real-time dense mapping with limited computation resource occupancy. We conducted extensive real-world experiments to validate the superior performance of OmniNxt in practical applications. All the hardware and software are open-access at\footnote[3]{\url{https://github.com/HKUST-Aerial-Robotics/OmniNxt}}, and we provide docker images of each crucial module in the proposed system. 
Project page: \urlstyle{same}\url{https://hkust-aerial-robotics.github.io/OmniNxt}.

\end{abstract}
\section{Introduction}
\label{sec:intro}

In recent years, remarkable advancements in open-source aerial robotics platforms have led to a surge in practical applications, \textit{e.g.}, exploration\cite{zhou2020fuel}, reconstruction\cite{feng2023fc}, and rescue\cite{lygouras2018rolfer}. However, it becomes evident that existing platforms, like FLA\cite{FLADrone}, MRS\cite{MRSDrone}, and Agilicious\cite{Agilicious}, are challenging for effectively performing these tasks in increasingly complex and dynamic environments due to their limited FoV.

Contrary to platforms with limited FoV, the early development of aerial robots with omnidirectional FoV by Gao \textit{et al.}\cite{AutonomousAerialRobotUsingDualfisheyeCameras} demonstrates the potential of omnidirectional FoV in performing the above tasks more robustly and efficiently. As proved by Zhang \textit{et al.}\cite{BenefitOfLargeFieldofView} and 
Wang \textit{et al.}\cite{wang2022lfvio}, the omnidirectional FoV improves the localization accuracy in challenging environments. Additionally, omnidirectional FoV enables minimizing the indirect-controlled yaw rotation\cite{Minimalsnap} during the flight\cite{Minimalsnap}, which enhances the energy efficiency of downstream tasks like autonomous navigation.

However, the lack of open-source omnidirectional FoV platforms can be attributed to the following challenges. \textbf{(1)} Sensor configuration: The standard way to achieve omnidirectional FoV \cite{AutonomousAerialRobotUsingDualfisheyeCameras} is using the multi-fisheye camera set. Unlike off-the-shelf camera modules, this set requires substantial efforts in structural design, hardware development, and meticulous calibration. \textbf{(2)} Algorithm development: To comprehensively leverage the benefits offered by the omnidirectional FoV, all algorithms, \textit{i.e.}, localization, mapping, planner, and controller should be adapted for properly exploiting the increased information. \textbf{(3)} System integration: Extensive testing and optimization of each component are essential to ensure stable and reliable performance within systems constrained by size and computational resources. System latency, resource occupancy, and overall performance should be carefully considered and optimized for robust and stable functionality in real-world scenarios.

To tackle these challenges and ensure platforms are broadly applicable to the research community, \textit{\textbf{COPE}} criteria have been distilled from existing works and anticipated future challenges to guide the platform design.
\begin{itemize}
    \item \textit{\textbf{C}ompact}: At the hardware level, sensor components should be compact to minimize the size and weight of the platform, thereby improving maneuverability. Additionally, critical software modules should consume minimal system resources, allowing a flexible environment for further tasks and development.
    \item \textit{\textbf{O}pen-source}: All hardware and software should be open-source, which facilitates cohesive reproduction, development, and enhancement by the community.
    \item \textit{\textbf{P}erceptive}: Comprehensive information about the surroundings and low-noise measurement should be perceived by sensors such as cameras and IMU. On the other hand, corresponding algorithms should be capable of effectively processing and utilizing the input from sensors to improve the system's robustness and stability.
    \item \textit{\textbf{E}xtendable}: The hardware should be developed to allow seamless interchange and migration across platforms, thus amplifying the utility of the present design. In parallel, the software should be modular and easily updatable to expedite the validation of emerging algorithms and adaptation for subsequent applications.
\end{itemize}

Following the \textit{\textbf{COPE}} criteria, we introduce OmniNxt, the first fully open-source aerial robotics platform with omnidirectional visual perception. OmniNxt boasts a compact size with exceptional computational resources. Thanks to our meticulous development, we endow OmniNxt with an expansive perception range and efficient resource consumption. Thorough evaluations demonstrate the effectiveness and advanced performance in real-world applications. The contributions of our system can be summarized as follows:
\begin{enumerate}

\item An open-source and compact hardware platform with omnidirectional FoV: We develop a coin-size yet high-performance flight controller Nxt-FC, which provides 500 Hz low-noise IMU data for the visual inertial odometry (VIO). In addition, we develop a multi-fisheye camera set to support omnidirectional perception. This set also offers synchronized images to VIO, where camera drivers and calibration tools are also provided.

\item An open-source and real-time omnidirectional perception framework: It consists of two principal components, Omni-VINS and Omni-Depth. The former provides accurate VIO, while the latter facilitates omnidirectional dense mapping. Both are devised as resource-efficient modules (See Fig. \ref{fig:system_overview}), which pave the way for advanced development in downstream tasks.
 
\item Extensive real-world experiments are conducted to evaluate the proposed system thoroughly. Experiments demonstrate that OmniNxt achieves outperforming localization accuracy and dependable point cloud quality. Moreover, autonomous navigation tests in a cluttered indoor environment validate the practicality of the proposed platform. Experiments also verify the adaptability of OmniNxt, highlighting its compatibility with a range of sensors and its ease of modification to suit diverse applications.

\end{enumerate}

\section{Related Works}

\label{sec:related_works}
\subsection{Omnidirectional Visual Inertial Odometry}
\label{sec:omnidirectional_vio}
\setcounter{figure}{1}
\begin{figure}[t]
    \vspace{0.05cm}
    \centering
    \includegraphics[width=0.95\linewidth]{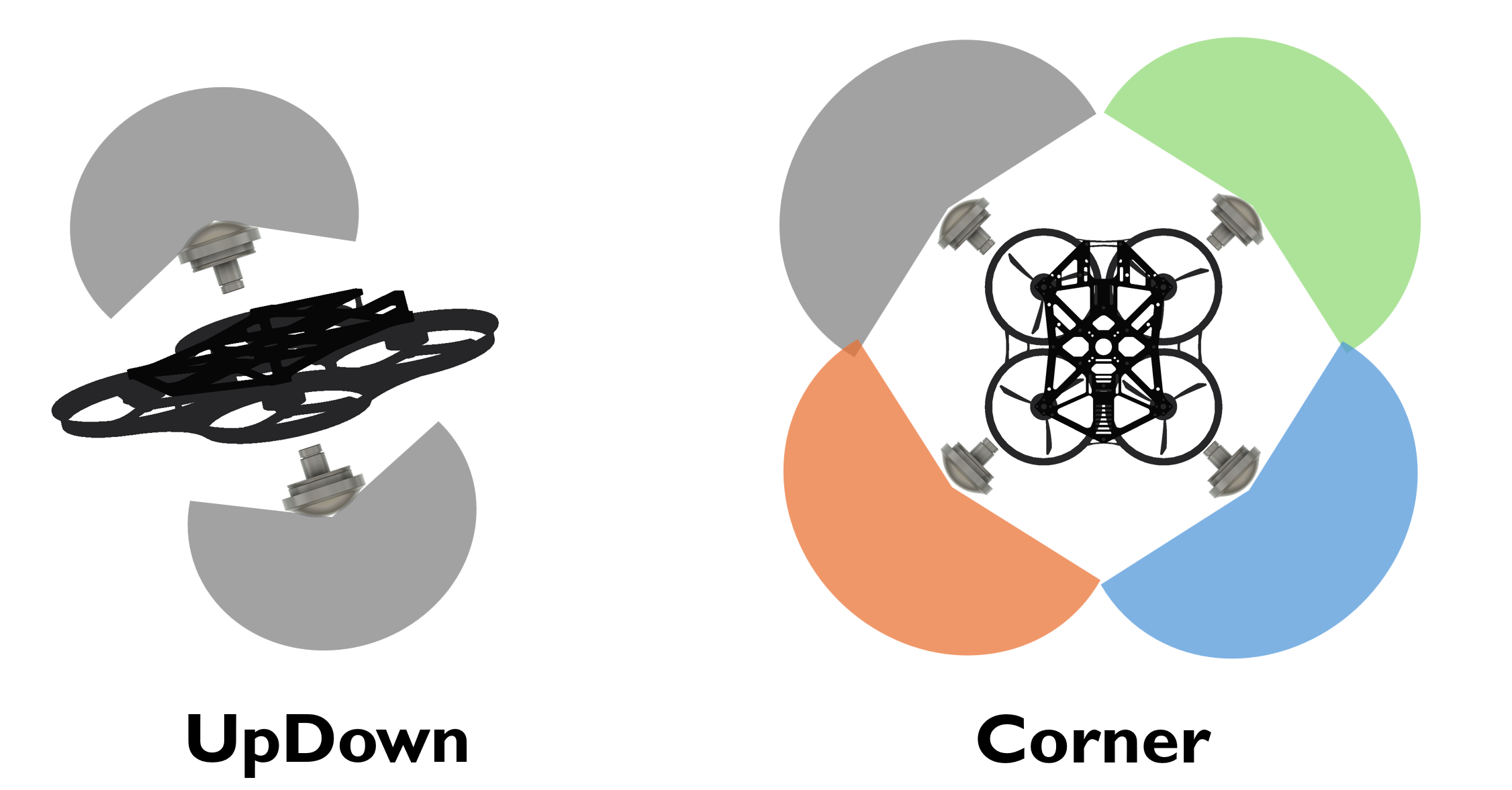}
    \caption{Typical structures of omnidirectional FoV. Cameras in the UpDown structure are on the top and bottom of the platform, facing upwards and downwards. The Corner structure places cameras on the corners of a plane, covering an omnidirectional view.}
    \label{fig:omni_structures}
    \vspace{-0.6cm}
\end{figure}

\begin{figure*}[t]
    \centering
    \vspace{0.05cm}
    \includegraphics[width=0.95\linewidth]{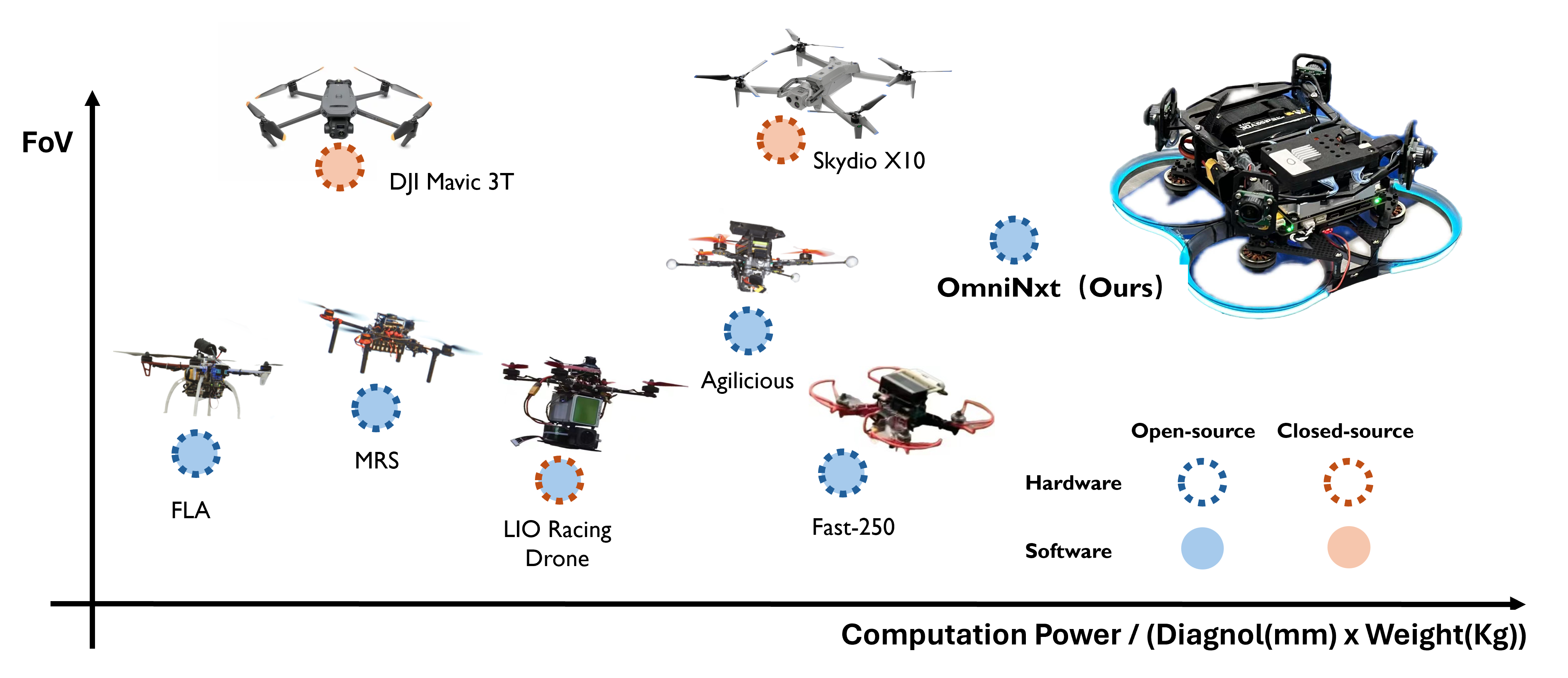}
    \vspace{-0.10cm}
    \caption{Platforms comparison. The platforms are compared based on their FoV and the ratio of onboard computation power to the product of size and weight. A higher value on the vertical axis indicates the platform's ability to perceive the surrounding environment more comprehensively. On the horizontal axis, a higher value represents a greater computational power available within a smaller size and weight, indicating a stronger capability of the platform to support downstream tasks.}
    \label{fig:platform_comparision}
    \vspace{-0.2cm}
\end{figure*}

\begin{table*}[t]
\begin{center}
    \vspace{0.2cm}
    \caption{A comparison of different available consumer and research platforms. We evaluate them in hardware dimensions, weight, sensor types, perception information types, perception range, and extendability. HW: hardware. SW: software.}
    \renewcommand\arraystretch{1.2}
    \begin{tabular}{c|c|c|c|c|c|ccc|c|c}
    \hline
    \specialrule{0em}{0.3pt}{0.3pt}
    \multirow{2}{*}{\textbf{Platforms}} & \multicolumn{2}{c|}{\textbf{Compact}} & \multicolumn{2}{c|}{\textbf{Open-source}} & \multicolumn{4}{c|}{\textbf{Perceptive}} & \multicolumn{2}{c}{\textbf{Extendable}} \\ \cline{2-11} 
     & \multicolumn{1}{c|}{HW} & SW & \multicolumn{1}{c|}{HW} & SW & \multicolumn{1}{c|}{Visual Info} & \multicolumn{1}{c|}{Geometry Info} & \multicolumn{1}{c|}{Sensor Type} & FoV & \multicolumn{1}{c|}{HW} & SW \\     \specialrule{0em}{0.3pt}{0.3pt} \hline \hline \specialrule{0em}{0.3pt}{0.3pt}
    {DJI MAVIC 3T\cite{mavic_3t}} &380mm \space 920g & - & \color{red}\XSolidBrush  & \color{red}\XSolidBrush  & \color{green}\Checkmark & \multicolumn{1}{c|}{\color{green}\Checkmark} & \multicolumn{1}{c|}{\begin{tabular}[c]{@{}c@{}} Multi-fisheye\\camera set\end{tabular}} & 360$^\circ$  & \color{red}\XSolidBrush & \color{red}\XSolidBrush \\      \specialrule{0em}{0.3pt}{0.3pt} \hline  \specialrule{0em}{0.3pt}{0.3pt}
     \href{https://www.skydio.com/x10}{Skydio X10\cite{skydio_x10}} & $\approx$380mm \space 2110g & - & \color{red}\XSolidBrush  & \color{red}\XSolidBrush & \color{green}\Checkmark & \multicolumn{1}{c|}{\color{green}\Checkmark} & \multicolumn{1}{c|}{\begin{tabular}[c]{@{}c@{}} Multi-fisheye\\camera set\end{tabular}} & 360$^\circ$ & \color{red}\XSolidBrush & \color{red}\XSolidBrush \\     \specialrule{0em}{0.3pt}{0.3pt} \hline \specialrule{0em}{0.3pt}{0.3pt}
  
    FLA\cite{FLADrone} & 450mm \space $\approx$2500g & Low & \color{green}\Checkmark &  \color{green}\Checkmark& \color{green}\Checkmark & \multicolumn{1}{c|}{\color{green}\Checkmark} & \multicolumn{1}{c|}{\begin{tabular}[c]{@{}c@{}}Camera \&\\ 2D-LiDAR\end{tabular}} & $\approx$90$^\circ$  & \color{green}\Checkmark  & \color{green}\Checkmark  \\     \specialrule{0em}{0.3pt}{0.3pt} \hline \specialrule{0em}{0.3pt}{0.3pt}
    MRS\cite{MRSDrone} & 450mm \space $\approx$1500g & Low & \color{green}\Checkmark & \color{green}\Checkmark & \color{green}\Checkmark & \multicolumn{1}{c|}{\color{green}\Checkmark} & \multicolumn{1}{c|}{\begin{tabular}[c]{@{}c@{}}Camera \&\\ 3D-LiDAR\end{tabular}} & $\approx$90$^\circ$  & \color{green}\Checkmark & \color{green}\Checkmark \\      \specialrule{0em}{0.3pt}{0.3pt}  \hline \specialrule{0em}{0.3pt}{0.3pt}
    \begin{tabular}[c]{@{}c@{}}LIO Racing \\ Drone\end{tabular}\cite{point_lio_drone} & 330mm \space $\approx$1300g & Median  & \color{red}\XSolidBrush & \color{red}\XSolidBrush  & \color{red}\XSolidBrush & \multicolumn{1}{c|}{\color{green}\Checkmark } & \multicolumn{1}{c|}{3D-LiDAR} & $\approx$70$^\circ$  & \color{red}\XSolidBrush & \color{green}\Checkmark  \\    \specialrule{0em}{0.3pt}{0.3pt} \hline \specialrule{0em}{0.3pt}{0.3pt}
    Agilicious\cite{Agilicious} &330mm \space $\approx$775g  & High  & \color{green}\Checkmark  & \color{green}\Checkmark  & \color{green}\Checkmark & \multicolumn{1}{c|}{\color{green}\Checkmark} & \multicolumn{1}{c|}{Stereo camera} & $\approx$90$^\circ$  & \color{red}\XSolidBrush & \color{green}\Checkmark  \\    \specialrule{0em}{0.3pt}{0.3pt} \hline \specialrule{0em}{0.3pt}{0.3pt}
     {Fast-250\cite{FAST_250}} &250mm \space $\approx$1000g  & Median & \color{green}\Checkmark  & \color{green}\Checkmark  & \color{green}\Checkmark  & \multicolumn{1}{c|}{\color{green}\Checkmark } & \multicolumn{1}{c|}{Stereo camera} & $\approx$90$^\circ$  & \color{red}\XSolidBrush & \color{green}\Checkmark \\   \specialrule{0em}{0.3pt}{0.3pt} \hline \hline \specialrule{0em}{0.3pt}{0.3pt}
    \textbf{OmniNxt(Ours)} & \textbf{88.92mm} \space \textbf{660g} & \textbf{High}  & \color{green}\Checkmark  & \color{green}\Checkmark & \color{green}\Checkmark & \multicolumn{1}{c|}{\color{green}\Checkmark} & \multicolumn{1}{c|}{\begin{tabular}[c]{@{}c@{}}Multi-fisheye\\camera set\end{tabular}} & \textbf{360$^\circ$} & \color{green}\Checkmark  & \color{green}\Checkmark \\     \specialrule{0em}{0.3pt}{0.3pt} \hline 
    \end{tabular}

    \label{tab:accessible_platforms}
    \vspace{-0.6cm}
\end{center}
\end{table*}

According to Zhang \textit{et al.}\cite{BenefitOfLargeFieldofView}, the FoV and the placement of the camera directly impact the odometry accuracy in different scenarios. An adequately installed camera with extended FoV enhances the effectiveness of feature extraction and tracking, contributing to improved odometry accuracy compared with limited FoV. To realize the omnidirectional FoV, existing works commonly adopt two typical configurations in structure: UpDown and Corner (Fig. \ref{fig:omni_structures}). In the UpDown structure, cameras usually feature 250$^\circ$ to 360$^\circ$ FoV, while the Corner structure usually utilizes three or more cameras with $180^\circ$ to $210^\circ$ FoV. Compared with the Corner structure, the UpDown structure suffers more from image distortions, while the Corner structure involves more complex calibration.

Corresponding to the evaluation in \cite{BenefitOfLargeFieldofView}, systems based on UpDown structure, such as the one proposed by Wang \textit{et al.} \cite{wang2022lfvio}, demonstrate higher odometry accuracy in indoor environments. This can be attributed to the greater probability of feature extraction and tracking in less distorted areas (center) of the images. However, their performance tends to be less favorable in outdoor settings. This discrepancy arises from the fact that features are more likely to be extracted and tracked in the greater distorted areas (margin) when the system performs outdoors, subsequently introducing more noise into the estimation of system status.

Works like Omnidirectional DSO \cite{OmniDSO}, based on Corner structure, achieve higher accuracy in both scenarios because of the less distorted image input. However, compared with the feature-based method like VINS\cite{qin2017vins}, the direct method is unsuitable for agile maneuvering platforms due to its sensitivity to motion blur and changes in image illumination.

\subsection{Omnidirectional Depth Estimation}
Retrieving dense depth information from multiple images captured by fisheye cameras is challenging due to the significant distortions. Existing solutions to this problem can be categorized into direct and indirect methods.

Most direct methods, such as OmniMVS\cite{OmniMVS}, employ neural networks to obtain dense depth information directly from the raw images. However, these methods often struggle with generalization across different cameras and can hardly meet the real-time requirement on resource-limited platforms.

On the other hand, the indirect methods aim to transfer the raw images into multiple pairs of stereo images captured by the virtual pinhole cameras (See Sec. \ref{subsec:omnidirectional_depth_estimation}).  Gao \textit{et al.}\cite{AutonomousAerialRobotUsingDualfisheyeCameras} leverage semi-global matching (SGBM) to obtain a dense depth estimation on resource-limited platforms. While Xie \textit{et al.}\cite{Xie_OmniVidar} employ a CNN to obtain a denser depth estimation.  Our approach is similar to \cite{Xie_OmniVidar}, but we achieve real-time inference speed on resource-limited platforms while maintaining a more flexible structure in realization (Sec. \ref{subsec:omnidirectional_depth_estimation}).


\subsection{Available Platforms}

Numerous research groups have made valuable contributions to the community by sharing their design. The design of these platforms varies greatly depending on the research topics they address. Based on the \textit{\textbf{COPE}} criteria, TABLE. \ref{tab:accessible_platforms} provides an overview of currently accessible platforms.

Existing platforms like FLA\cite{FLADrone} and MRS\cite{MRSDrone} are designed for data collection and autonomous navigation in open areas. These platforms share a similar solution, utilizing a PX4\cite{Meier2015PX4AN} based flight controller, a CPU-only onboard computer, mono or stereo cameras, and LiDAR. However, achieving compactness and lightweight in their design is challenging due to the introduction of LiDAR to obtain dense map. On the other hand, the platform, which is designed for aggressive planning and control, prioritizes maximizing the thrust weight ratio (TWR). One notable platform in this category is Agilicious\cite{Agilicious}. Different from those platforms mentioned above, Agilicous\cite{Agilicious} achieves compactness and lightweight by adopting a cramped hardware design, which allows the platform to have high TWR but makes it hard to integrate additional sensors. This limitation poses challenges in accessing a comprehensive surrounding environment, limiting its applicability to future tasks. Besides, commercial platforms like DJI MAVIC 3T\cite{mavic_3t} and SKYDIO X10\cite{skydio_x10} are not open-source, hindering further development and algorithm validation. Their large dimensions and weight also limit the range of applications.

\section{OmniNxt Platform}
\label{sec:omninxt_platform}
\subsection{System Overview}
OmniNxt is designed to be a general open-source aerial robotics platform. Therefore, we strictly follow the \textbf{COPE} criteria in hardware and software design.

As shown in Fig. \ref{fig:system_overview}, OmniNxt consists of three key components in hardware: First, We use Nvidia Jetson Orin NX for the onboard computer, which has an 8-core CPU running at 2.0 GHz and a GPU with 1024 CUDA cores. CPU and GPU share the unified 16G RAM. Second, we adopt a multi-fisheye camera set to capture the omnidirectional image at 20 Hz. All four cameras are synchronized by one camera triggering the other three. Last, to further enhance the robustness of OmniNxt, we develop a low-noise flight controller, which provides 500 Hz low-noise IMU data. Besides, the software framework of OmniNxt can be divided into the following four parts, each serving a specific function:
\begin{itemize}
    \item \textbf{Omni-VINS}: This module combines the high-frequency IMU measurements (500 Hz) and the omnidirectional image (20 Hz) to generate accurate VIO, which is crucial for mapping, planning, and control modules. (Sec. \ref{subsec:omnidirectional_vins})
    \item \textbf{Omni-Depth}: This module performs real-time omnidirectional dense point cloud generation by processing the omnidirectional image with its virtual-stereo frontend and multi-stream-inference backend. (Sec. \ref{subsec:omnidirectional_depth_estimation})
    \item \textbf{Planner}:  Based on the Omni-VINS and Omni-Depth, this module generates the trajectory toward the goal position following the aerial robotic dynamics. (Sec. \ref{subsec:planner_and_controller})
    \item \textbf{Controller}: The trajectory generated from the planner is converted to a low-level attitude command by this module and then executed by the flight controller. (Sec. \ref{subsec:planner_and_controller})
\end{itemize}
The perception modules (Omni-VINS, Omni-Depth, and related hardware drivers) in OminNxt occupy 27\% CPU and 40\% GPU onboard resources, sparing sufficient resources for the downstream tasks.

\subsection{Nxt-FC}
\label{subsec:flight_controller}
 The lack of open-source hardware options in most available flight controllers presents a significant challenge in further development and adaptation on various platforms. To address this issue, we design and open-source a flight controller named Nxt-FC with compact dimensions of 27mm$\times$33mm. Our flight controller is based on open-source autonomous pilot firmware PX4\cite{Meier2015PX4AN}. We develop a high-frequency raw IMU data stream for robust and accurate VIO.  Detailed information is available on our \href{https://hkust-aerial-robotics.github.io/OmniNxt}{project page}.

\subsection{Multi-fisheye Camera Set Calibration}
\label{subsec:multi_fisheye_camera_calibration}

\begin{figure}[t]
    \centering
    \vspace{0.15cm}
    \includegraphics[width=0.93\linewidth]{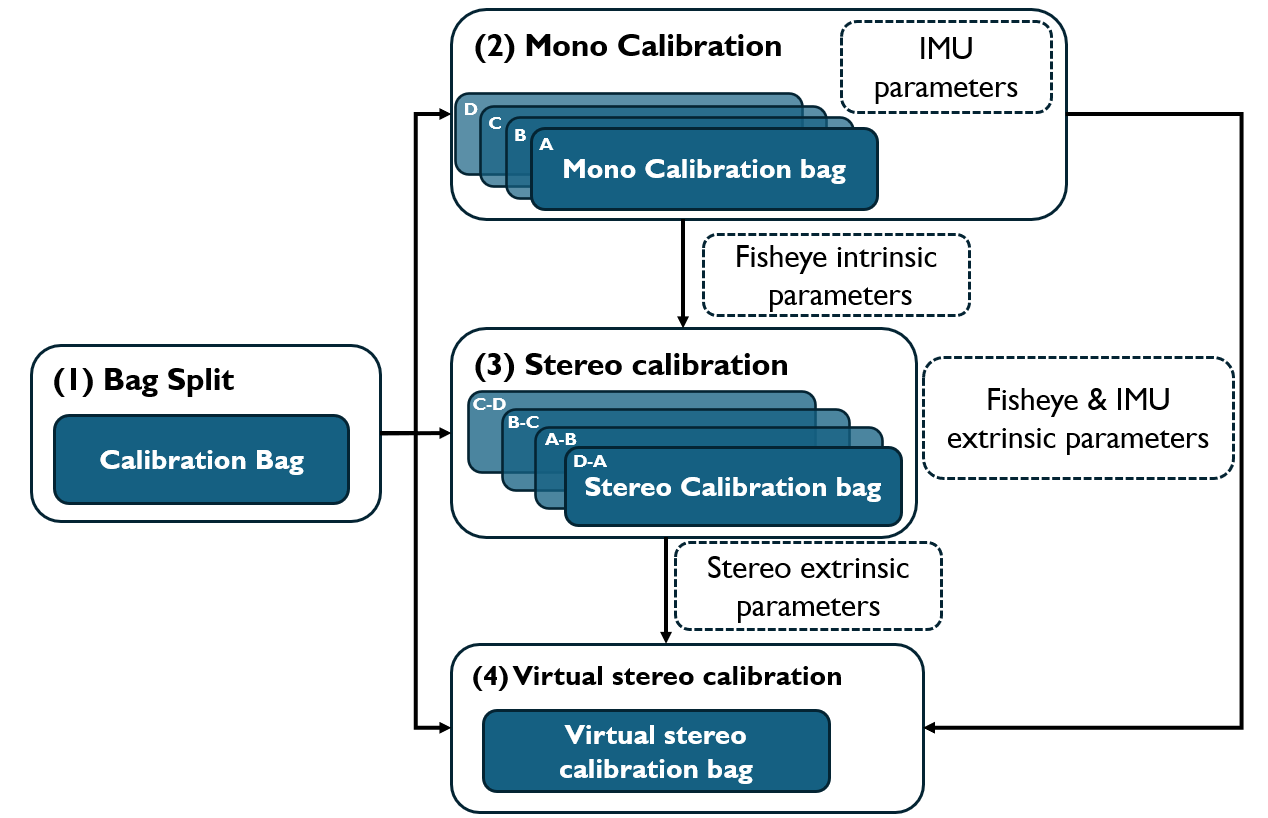}
    \caption{The calibration pipeline of multi-fisheye camera set. The numbers indicate the calibration sequence. The letter on the top left of each box corresponds to the camera index in Fig. \ref{fig:virtual_stereo}.B}
    \label{fig:calibration_pipline}
    \vspace{-0.6cm}
\end{figure}

Calibrating all four fisheye cameras' intrinsic and extrinsic parameters together results in a complex optimization problem, thus increasing processing time and lowering the success rate. 
To address this issue, our pipeline utilizes tartankalibr\cite{duisterhof2022tartancalib} in fisheye cameras calibrations and kalibr-toolbox\cite{unified_multi_sensor}\cite{extend_kalibr} in virtual pinhole cameras calibrations. The calibration process is illustrated in Fig. \ref{fig:calibration_pipline}.

Initially, we calibrate the intrinsic parameters of each fisheye camera. Once the projection error of each camera is less than 0.5 pixels, we proceed to calibrate the extrinsic parameters between adjacent cameras as well as between cameras and IMU.

After having established the intrinsic and extrinsic parameters of the multi-fisheye camera set, we undistort the image based on the cylindrical camera model (Sec. \ref{subsec:omnidirectional_vins}) to generate virtual stereo pairs (Fig. \ref{fig:virtual_stereo}.C) and calibrate both intrinsic and extrinsic parameters of virtual stereo cameras. The transformation between fisheye and virtual stereo cameras is further explained in Sec. \ref{subsec:omnidirectional_depth_estimation}.

Our pipeline reduces the calibration time by concurring the calibration process and raises the success rate by allowing the calibration process to resume from the failed case quickly.

\subsection{Omni-VINS}
\label{subsec:omnidirectional_vins}
Omni-VINS is optimized for onboard real-time performance based on our previous work $D^2$SLAM\cite{xu2023d2slam}. The significant distortion in fisheye images poses challenges in feature extraction and feature tracking, particularly when using CNN-based methods that are trained with images gathered by the pinhole camera. To address this issue, we employ the MEI model\cite{mei_model} for the intrinsic parameters of raw fisheye cameras and radial-tangential models to formulate the distortion. First, We undistort the fisheye camera image with the cylindrical camera model. The cylindrical camera model\cite{plaut20213d} can be written as:
\begin{equation}
\begin{aligned}
    \begin{bmatrix}
    u\\
    v\\
    1
    \end{bmatrix} &= \begin{bmatrix}
    f_\phi  & 0 & u_0 \\
    0  & f_y & v_0\\
    0  & 0 & 1 
    \end{bmatrix}\begin{bmatrix}
     \phi\\
     Y/\rho\\
    1
    \end{bmatrix}, \\
    \phi &= atan2(X,Z),\\
    \rho &= \sqrt{X^2+Z^2},
\end{aligned}
\end{equation}
where $f_\phi$ and $f_y$ are the focal length. This model offers the advantage of representing virtual cameras with adjustable FoV and rotations relative to the original camera.

In our implementation, we set $f_\phi$ to $\frac{190^{\circ}}{W}$ ($W$ is the width of the image) to eliminate the unexposed margin of the image. To ensure the consistency of the undistorted image, we set $f_y$ equal to $f_\phi$.

We then perform feature extraction and tracking based on the cylindrical undistorted image. Omni-VINS heritages the hybrid feature extraction strategy developed in $D^2$VINS\cite{xu2023d2slam} but changes the tracking method from Superglue\cite{sarlin2020superglue} to  Lucas-Kanade (LK)\cite{LKinproceedings} method to minimize the GPU occupancy and accelerates the matching process. 
We perform feature tracking in both previous and latest frames and among adjacent cameras' frames.

Our previous works \cite{qin2017vins} and \cite{AutonomousAerialRobotUsingDualfisheyeCameras} have thoroughly formulated the visual-inertial problem. The full state vector in the sliding window is defined as:
\begin{equation}
\begin{aligned}
\mathcal{X} &= [x_{0},x_{1},..., x_{m},x_{C_{0}}^{b},x_{C_{1}}^{b},...,x_{C_{n}}^{b},\lambda_{0},\lambda_{1},..., \lambda_{l}], \\
x_{i} &= [p_{b_{i}}^{W},v_{b_{i}}^{W},q_{b_{i}}^{W},b_{a}^{b},b_{g}^{b}], k \in [0,m], \\
x_{c_{j}}^{b} &= [p_{c_{j}}^{b},q_{c_{j}}^{b}], j \in [0,n],
\end{aligned}
\end{equation}
where $m$ is the total number of keyframes, $\lambda_{i}$ is the inverse depth of the $i^{th}$ feature from its first observation, and $n$ is the number of fisheye cameras.
The visual inertial odometry problem in Omni-$D^2$VINS is defined as:
\vspace{-0.2cm}
\begin{align}
		&\min_{\mathcal{X}}{\Biggl \{ {{\left \|r_{p}-H_{p}\mathcal{X}\right\|}^{2}}  + \sum_{K \in \mathcal{B}}{ {\left \| r_{\mathcal{B}}({\hat{z}}^{b_{k}}_{b_{k+1}},\mathcal{X}) \right \|}^{2}_{P^{b_k}_{b_{k+1}}}} + } \nonumber \\
		&{\sum_{(l,j) \in \mathcal{C}}{{\left \| r_{\mathcal{C}}(\hat{z}^{c_{j}}_{l},\mathcal{X}) \right \|}^{2}_{P^{c_{j}}_{l}}}  \Biggr \} }
\end{align}
We use Ceres Solver\cite{Agarwal_Ceres_Solver_2022} for solving this least squares problem.
\vspace{-0.1cm}

\subsection{Omni-Depth}
\label{subsec:omnidirectional_depth_estimation}
Omni-Depth is designed with a modular structure consisting of a virtual-stereo frontend and a multi-stream-inference backend to ensure the generalization in various stereo-matching techniques.

\begin{figure}[t]
    \centering
    \vspace{0.15cm}
    \includegraphics[width=0.95\linewidth]{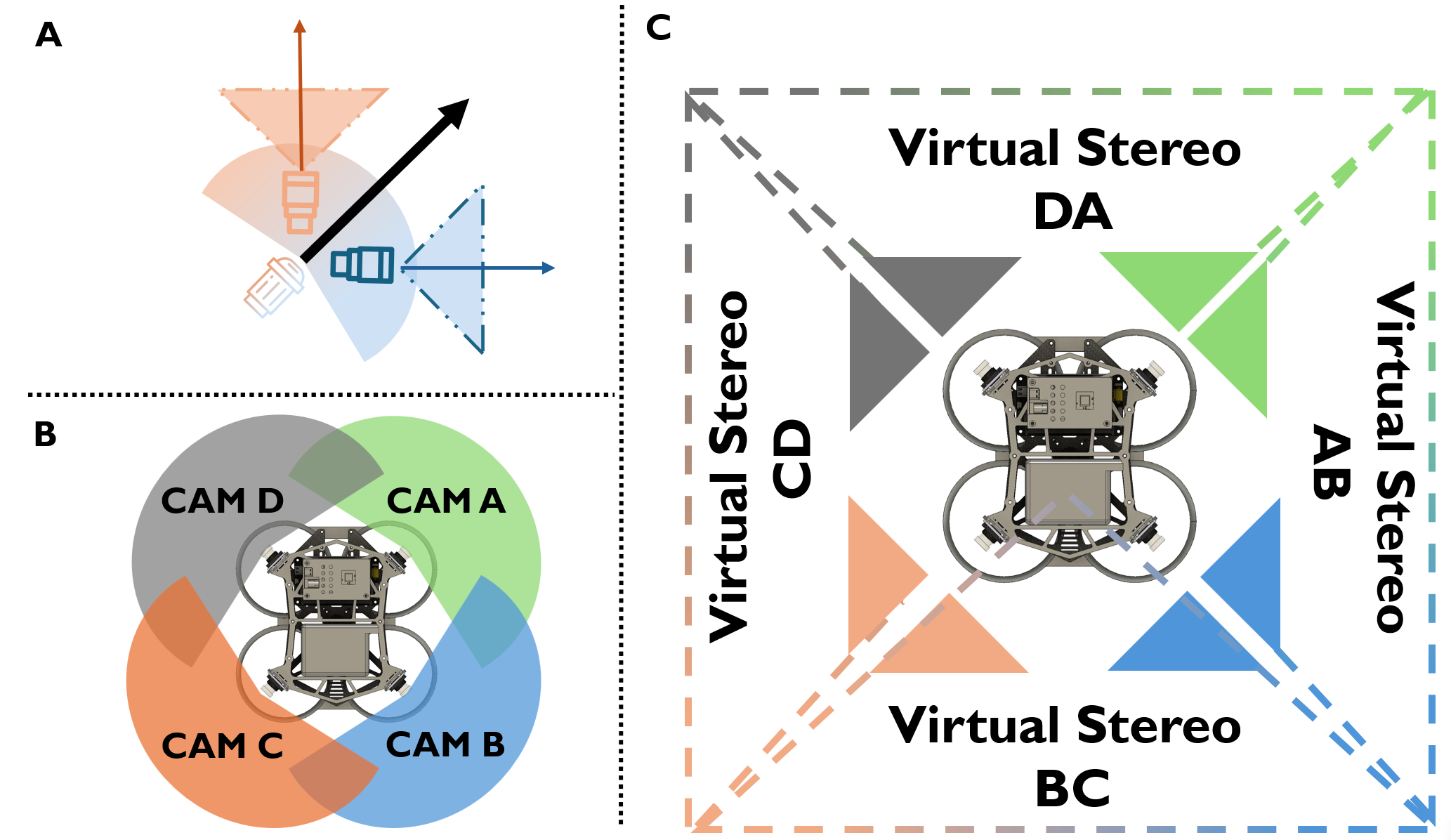}
    \caption{Illustration of virtual-stereo frontend. \textbf{A}: the Z axis of virtual cameras (Orange and Blue) and the fisheye camera (Black). \textbf{B}: the placement of four fisheye cameras. \textbf{C}: shows the virtual stereo pairs.}
    \vspace{-0.6cm}
    \label{fig:virtual_stereo}
\end{figure}

\begin{figure*}[t]
    \centering
    \vspace{0.4cm}
    \includegraphics[width=1.0\linewidth]{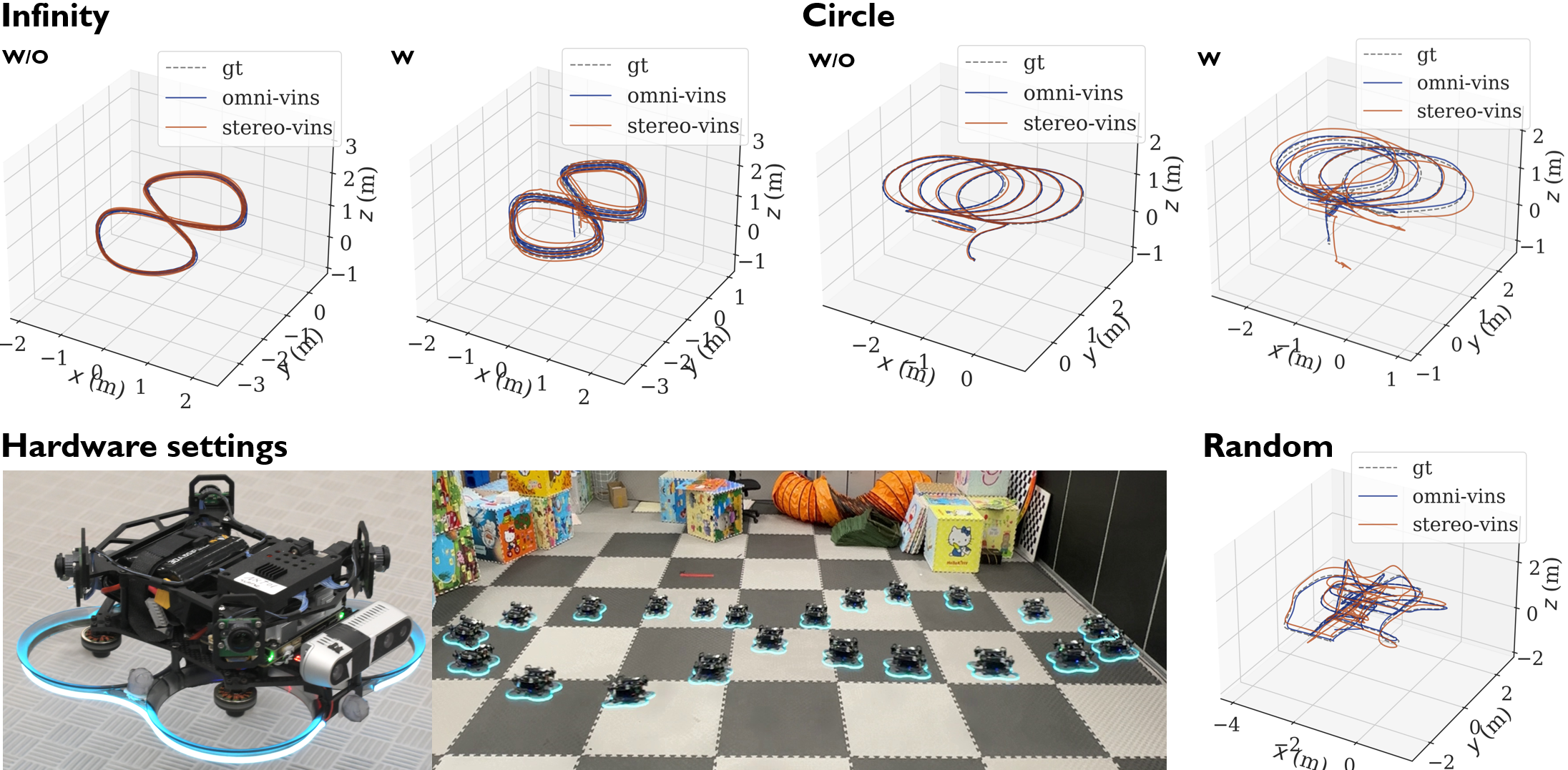}
    \caption{Omni-VINS evaluation. \textbf{Infinity}, \textbf{Circle}, and \textbf{Random} are the three testing trajectories. \textbf{W/O} means flight without rotating the yaw to the speed direction. \textbf{gt} is the ground truth. \textbf{W} means flight while the yaw direction follows the speed direction. The data collection platform and testing environment are shown in \textbf{Hardware settings}. (Sec. \ref{subsec:vins_evaluation})}
    \label{fig:vins_comparison}
    \vspace{-0.4cm}
\end{figure*}

The virtual-stereo frontend is demonstrated in Fig. \ref{fig:virtual_stereo}.C, where the fisheye camera images are undistorted into two perpendicularly placed virtual cylindrical cameras. The FoV of virtual cameras is set to $100^\circ$. The extrinsic parameters of the virtual pinhole cameras can be defined as follows:
\begin{equation}
\begin{aligned}
    &\mathbf{T}_{vcam} = 
    \begin{bmatrix}
     \mathbf{R}_{fisheye}^{vcam} & 0 \\
      0&1
    \end{bmatrix}
    \begin{bmatrix}
     \mathbf{R}_{fisheye} & \mathbf{t}_{fisheye} \\
      0&1
    \end{bmatrix} ,  \\
    &\mathbf{R}_{fisheye}^{vcam_{l}} = \mathbf{R}_{y}(-\frac{1}{4}\pi), \\ 
    &{\mathbf{R}_{fisheye}^{vcam_{r}}} = \mathbf{R}_{y}(\frac{1}{4}\pi), \\
    \end{aligned}
\end{equation}
where $vcam_l$ represents the left virtual camera, while $vcam_r$ represents the right one. Since the FoV of these two virtual cameras is close to the standard pinhole camera, we can calibrate them with the pinhole camera model.

In the backend of Omni-Depth, we employ CNN to estimate the disparity of each virtual stereo pair because of its more robust performance in the textureless environment. The Omni-Depth is flexible in accommodating different CNN models in the backend. This allows easy adaptation to meet various platforms' accuracy and inference speed requirements. See Sec. \ref{subsec:depth_estmiation} for the implementation details.

\subsection{Planner and Controller}
\label{subsec:planner_and_controller}
 Combining precise and reliable odometry from Omni-VINS and dense point cloud from Omni-Depth, OmniNxt offers extensive support for various tasks. Benefited from The omnidirectional perception, yaw rotation can be minimized during the flight, which simplifies the scheme of the trajectory planning in autonomous navigation (Sec. \ref{subsec:planning_evaluation} ) while increasing the accuracy of VIO (Sec. \ref{subsec:vins_evaluation}).
The command from the tasks level is fed into the controller module and then converted into the desired attitude command in MAVROS format. For efficient control, We utilize the open-source Px4-Controller\cite{FAST_250}.

\section{Experiments}
\label{sec:experiments}
We conduct a comprehensive analysis of OmniNxt's performance in VIO accuracy (Sec. \ref{subsec:vins_evaluation}) and omnidirectional dense map quality (Sec. \ref{subsec:depth_estmiation}) in the real world. In Sec. \ref{subsec:planning_evaluation}, we demonstrate OmniNxt's ability to autonomously navigate in a narrow indoor environment. In Sec. \ref{subsec:extension_of_omniNxt}, we also adapt OmniNxt with other common visual sensors to ensure its compatibility.

\begin{figure}[t]
    \centering
    \vspace{-0.05cm}
    \includegraphics[width=0.95\linewidth]{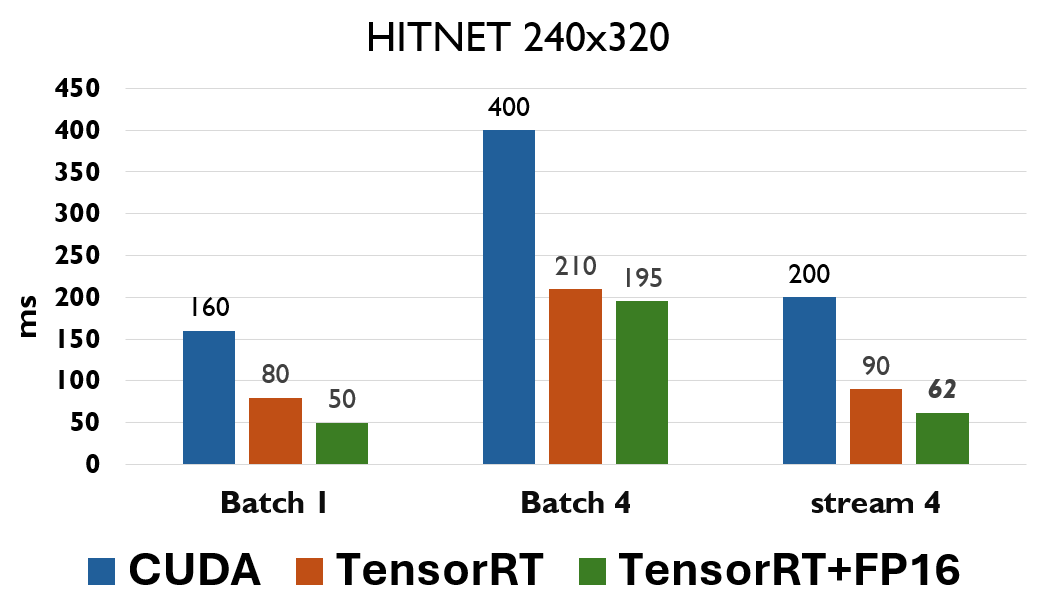}
    \caption{HITNET inference speed. Batch 1: The inference time of one group of data. Batch 4: The inference time of four groups of data concatenated on batch dimension. Stream 4: The inference time of four groups of data in multi-stream.}
    \label{fig:inference_speed}
    \vspace{-0.6cm}
\end{figure}

\begin{figure*}
    \vspace{0.25cm}
    \centering
    \includegraphics[width=0.99\linewidth]{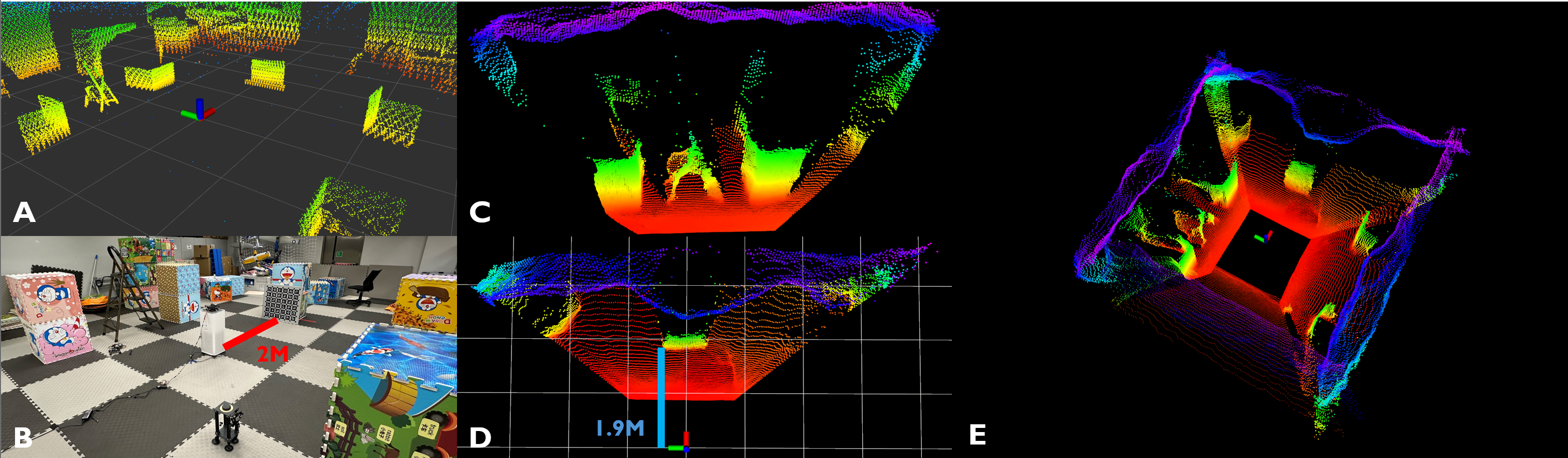}
    \caption{Omni-Depth evaluation. \textbf{A}: the point cloud collected by LIVOX Mid360. \textbf{B}: the testing environment. \textbf{C and D}: the detailed point cloud generated from the virtual stereo pair DA and CD (grid is 1m) as illustrated in Fig. \ref{fig:virtual_stereo}.C. \textbf{E}: The overview of the point cloud generated from Omni-Depth. (Sec. \ref{subsec:depth_estmiation})}
    \label{fig:pointcloud_overview}
\end{figure*}

\begin{figure*}
    \centering
    \includegraphics[width=0.99\linewidth]{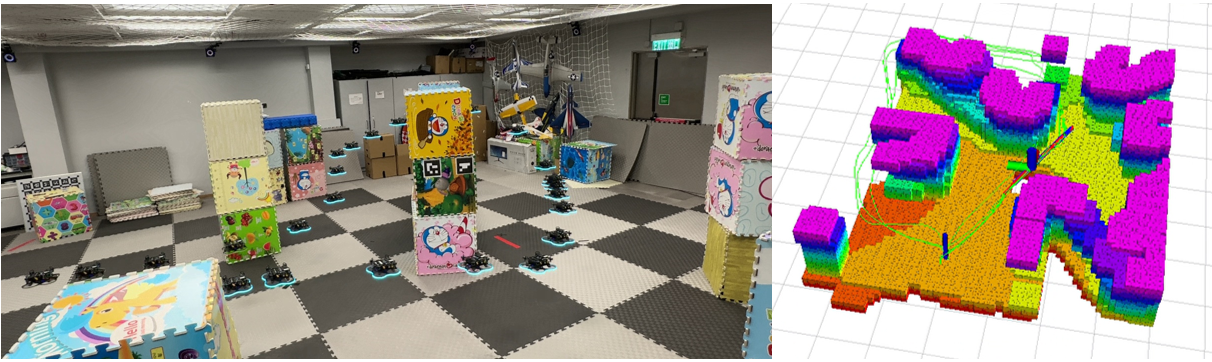}
    \caption{Real-world autonomous navigation tests. \textbf{Left}: OmniNxt autonomously flies without rotating the yaw to the speed direction. \textbf{Right}: the executed trajectory and voxel map. (Sec. \ref{subsec:planning_evaluation})}
    \label{fig:ego-plnner}
    \vspace{-0.4cm}
\end{figure*}

\subsection{Omni-VINS Evaluation}
\label{subsec:vins_evaluation}
We respectively equip OmniNxt with the multi-fisheye camera set and Intel D435\cite{D435} stereo camera to compare the VIO accuracy of omnidirectional VINS (Omni-VINS) and limited-FoV VINS (Stereo-VINS). The ground truth is collected by Opti-Track. In this evaluation, we design three trajectories: \textbf{Infinity}, \textbf{Cirecle}, and \textbf{Random}. In \textbf{Infinity} and \textbf{Cirecle} cases, OmniNxt follows the trajectory in two manners. One moves without rotating the yaw to the speed direction, and another moves while the yaw follows the speed direction. In the \textbf{Random} case, the human pilot controls the robots, randomly performing dashing, spinning, and hovering. Our test environment poses significant challenges to VIO because of the black wall on the right. The testing trajectories and environment are illustrated in Fig. \ref{fig:vins_comparison}. 

Omni-VINS and Stereo-VINS are configured with the same feature extraction and tracking numbers. The sliding window size is set to be 10. Both use good-feature-to-track features and the LK method for tracking. The extrinsic parameters of the camera and IMU in Stereo-VINS are calibrated by Kalibr\cite{extend_kalibr}.

Omni-VINS benefits from the omnidirectional FoV, achieving more robust and accurate VIO in all cases. For the root mean squared error (RMSE) of absolute trajectory error (ATE), see TABLE. \ref{tab:ATE_VINS_oratation} (Our evaluate method and tools based on EVO\cite{grupp2017evo}).

\begin{table}[h]
\vspace{-0.1cm}
\centering
\caption{Localization accuracy comparison (metric: RMSE).}
\begin{tabular}{l|c|c|c}
\hline
 & \multicolumn{1}{l|}{\textbf{Omni-VINS}} & \multicolumn{1}{l|}{\textbf{Stereo-VINS}} & \multicolumn{1}{l}{\textbf{Trajectory Length}} \\ \hline \hline
Infinity(W) & \textbf{0.084m} & 0.110m & 70.1m \\ 
Infinity(W/O) & \textbf{0.043m} & 0.052m & 70.1m \\ \hline
Circle(W) & \textbf{0.086m} & 0.268m & 60.2m \\
Circle(W/O) & \textbf{0.025m} & 0.050m & 43.0m \\ \hline
Random(W) & \textbf{0.098m} & 0.460m & 60.0m \\ \hline
\end{tabular}
\label{tab:ATE_VINS_oratation}
\vspace{-0.2 cm}
\end{table}

\subsection{Real-time Depth Estimation}
\label{subsec:depth_estmiation}
\vspace{-0.1cm}
In practice, we use the pre-trained HITNET\cite{HITNET} model with the input dimension (BWHC) 1$\times$320$\times$240$\times$2 in the backend to balance the estimation quality and inference speed. We accelerate the inference by utilizing Tensor Core and FP16 quantization. The inference time of one virtual stereo 
drops from 160 ms to 50 ms. Since Omni-Depth involves four sets of inferences for each frame, the sequential execution of these inferences last 200ms. To address this, we explore two approaches to parallelize the inference procedure. Firstly, we concatenate the four groups of input data along the batch dimension, resulting in an input dimension of 4$\times$320$\times$240$\times$2. However, this approach does not improve the inference time.
Then, we adopt NVIDIA multi-stream, which allows four sets of inference to be executed concurrently. Subsequently, the total inference time decreased from 200 ms to 62 ms. Fig. \ref{fig:inference_speed} demonstrates the inference speed of HINTET in different manners.
The Omni-Depth can run at 15 Hz onboard, providing a real-time dense map.  The bias of the dense map is around 10 cm. (See Fig. \ref{fig:pointcloud_overview}E)

\subsection{Real-world Autonomous Navigation Tests}
\label{subsec:planning_evaluation}
To demonstrate the practicability of OmniNxt, we conduct autonomous navigation tests in a cluttered indoor environment. Specifically, we modify the trajectory generation strategy in ego-planner\cite{zhou2020egoplanner} by removing the yaw angle trajectory optimization thanks to the omnidirectional perception. This modification enhances the accuracy of localization (as shown in Sec.\ref{subsec:vins_evaluation}) and improves the quality of the dense map by reducing the image blur caused by yaw rotation. This improvement ultimately enhances the efficiency and safety of trajectory planning. For the experiment, we set the maximum speed to 1.0 m/s and the maximum acceleration to 0.6 m/s$^2$. We represent the surrounding environments with volumetric mapping. All the modules are running onboard. (See Fig. \ref{fig:ego-plnner})

\subsection{Adaptation of OmniNxt}
\label{subsec:extension_of_omniNxt}
\begin{figure}[h]
    \centering
    \includegraphics[width=1\linewidth]{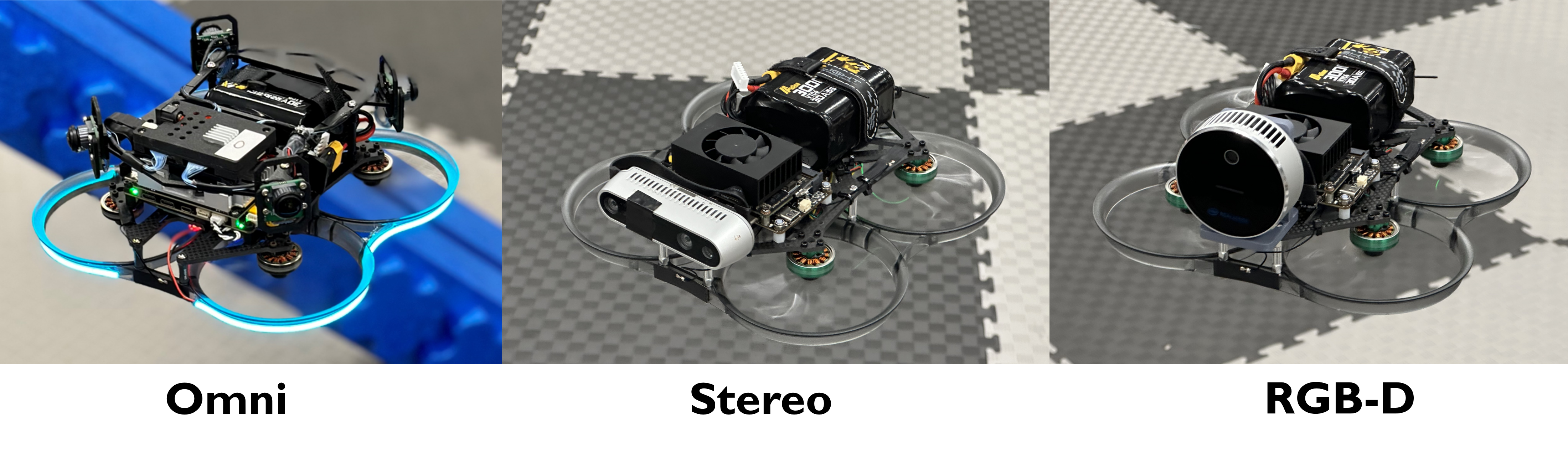}
    \caption{OmniNxt carries different types of visual sensors. Here, we demonstrate the installation of omnidirectional, stereo, and RGB-D cameras. }
    \label{fig:omni_platforms}
    \vspace{-0.4cm}
\end{figure}

OmniNxt boasts sophisticated hardware designs that facilitate the swift interchange of visual sensors. Presently, we offer two additional versions of sensor configurations: \textbf{Stereo} (Intel D435\cite{D435}) and \textbf{RGB-D} (Intel L515\cite{L515}), as depicted in Fig. \ref{fig:omni_platforms}. We have conducted real-world flight experiments to demonstrate their practicality (Stereo in Sec. \ref{subsec:vins_evaluation}, and RGB-D in H{$_2$}-Mapping\cite{H2mapping}). Detailed information is available on our \href{https://hkust-aerial-robotics.github.io/OmniNxt}{project page}.

\vspace{-0.1cm}
\section{Conclusion and Future Work}

In this paper, we present OmniNxt, a fully open-source and compact aerial robotics platform with omnidirectional visual perception. OmniNxt demonstrates exceptional performance in localization and dense mapping, with limited size and onboard computational resources. We develop Nxt-FC, a coin-size yet high-performance flight controller, which is general to the community. Besides, a multi-fisheye camera set is designed to support omnidirectional perception.
Based on our devised hardware, we also propose a real-time omnidirectional perception framework, including Omni-VINS and Omni-Depth.
Comprehensive real-world experiments demonstrated the superiority and practicability of OmniNxt. 
In the future, we aim to improve the capability of omnidirectional visual perception to enable exceptional performance in increasingly demanding environments.










\bibliography{references}

\end{document}